\def\year{2018}\relax
\documentclass[letterpaper,twocolumn]{article} %DO NOT CHANGE THIS
\usepackage{aaai18}  %Required
\usepackage{times}  %Required
\usepackage{helvet}  %Required
\usepackage{courier}  %Required
\usepackage{url}  %Required
\usepackage{graphicx}  %Required
\def\vector#1{\mbox{\boldmath $#1$}}
\usepackage{amsmath}
\usepackage{amsfonts}
\usepackage{booktabs}
\usepackage{color}
\usepackage{url}
\usepackage[implicit=false, colorlinks=true, urlcolor=blue]{hyperref}

\title{Hierarchical Video Generation from Orthogonal \\ Information: Optical Flow and Texture}
\author{Katsunori Ohnishi\thanks{indicates equal contribution.} \\
The University of Tokyo\\
ohnishi@mi.t.u-tokyo.ac.jp
\And 
Shohei Yamamoto\footnotemark[1] \\
The University of Tokyo \\
yamamoto@mi.t.u-tokyo.ac.jp
\And 
Yoshitaka Ushiku \\
The University of Tokyo\\
ushiku@mi.t.u-tokyo.ac.jp
\And 
Tatsuya Harada\\
The University of Tokyo / RIKEN\\
harada@mi.t.u-tokyo.ac.jp
}

\begin{document}
\maketitle

\begin{abstract}
Learning to represent and generate videos from unlabeled data is a very challenging problem. To generate realistic videos, it is important not only to ensure that the appearance of each frame is real, but also to ensure the plausibility of a video motion and consistency of a video appearance in the time direction. The process of video generation should be divided according to these intrinsic difficulties. In this study, we focus on the motion and appearance information as two important orthogonal components of a video, and propose Flow-and-Texture-Generative Adversarial Networks (FTGAN) consisting of FlowGAN and TextureGAN. 
In order to avoid a huge annotation cost, we have to explore a way to learn from unlabeled data. 
Thus, we employ optical flow as motion information to generate videos. FlowGAN generates optical flow, which contains only the edge and motion of the videos to be begerated. On the other hand, TextureGAN specializes in giving a texture to optical flow generated by FlowGAN. This hierarchical approach brings more realistic videos with plausible motion and appearance consistency. 
Our experiments show that our model generates more plausible motion videos and also achieves significantly improved performance for unsupervised action classification in comparison to previous GAN works. In addition, because our model generates videos from two independent information, our model can generate new combinations of motion and attribute that are not seen in training data, such as a video in which a person is doing sit-up in a baseball ground.
\end{abstract}

\section{Introduction}
\vspace{-1mm}
%%%動画に関する一般的なお話
Video understanding is a core problem in computer vision. Given the considerable progress in video recognition (e.g., action classification, event detection), video generation and unsupervised learning of video representation (e.g., future frame prediction) have been gaining considerable attention. Automatic video generation can potentially help human designers and developers to convert their high-level concepts into pixel-level videos. Some works \cite{vondrick2016generating,saito2016temporal} have tried to generate videos with generative adversarial networks (GANs) \cite{goodfellow2014generative} approach. 

%%%動画生成の難しさ
However, video generation is a very challenging task. The difficulties lie not only on that (a) each frame should be a realistic image as generated image, but also that (b) the same scene and foreground should be generated in each video and that (c) the generated video should have plausible motion. 

%%%手法概要
In this study, we focus on the above-mentioned fundamental difficulties of generating a realistic video and propose the generation of videos through GAN spitting into two orthogonal information types: motion and appearance. Because previous GANs for video \cite{vondrick2016generating,saito2016temporal} have not focused on these difficulties, they can not produce motion realistic videos.

Unsupervised video generation should learn without any annotation. Thus, we utilize optical flow as motion information. Calculating optical flow does not require annotations on each dataset and can be generally obtained with unsupervised method, which means our method still can be trained in an unsupervised manner. 

Our model, Flow and Texture Generative Adversarial Networks (FTGAN), consists of two GANs: FlowGAN and TextureGAN. 
We first generate optical flow with FlowGAN, and then convert optical flow into RGB videos with TextureGAN. This hierarchical approach is explained in detail below. 
%%stage1
The characteristics of optical flow are as follows: it has edges of moving objects, contains time-directional continuity, and does not contain texture information. Therefore, optical flow generation is much easier than RGB video generation. Optical flow generation first assures and achieves reasonable motion and rough edges for generated videos. 
%%stage2
Next, TextureGAN colors the optical flow supplementing rough outlines. TextureGAN provides texture information to the generated optical flow while maintaining scene and foreground consistency. Thus, we can obtain more realistic videos containing plausible motion. %\textcolor{blue}{As explained above, we factorize and simplify a video generative model into two complemental generative adversarial models.} 
%Because our model do not use any annotated information, it can be trained not only human action video but also on any other video dataset. 

\begin{figure}[t]
\begin{center}
  	\includegraphics[width=\hsize]{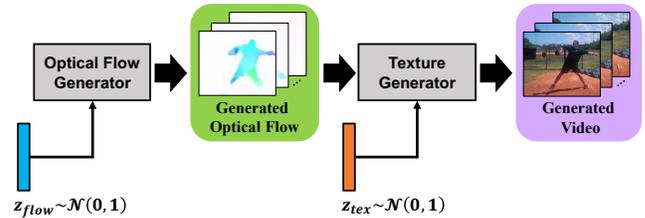}
\end{center}
\vspace{-7mm}
   \caption{Generative pipeline of our FTGAN.}
\vspace{-5mm}
   \label{fig:system_overview}
\end{figure}

%%%手法の面白い副産物
As an application for video generation, it is important for a model to express a wide range of datasets. However, in the existing video generation, the appeance and motion are limited to the combinations that exist in the dataset. 
Since motion and appearance information are learnt separately, our model can control the appearance and motion information of a generated video independently. 
Thus, our model can generate nonexisting motion and attribute combination in the training dataset comprising a video in which a person is doing sit-up in a baseball ground.

It is also known that GAN can be used as an unsupervised feature extractor. Thus, as well as previous generative adversarial networks for video, we also perform motion recognition experiments. On the action recognition dataset UCF101 \cite{soomro2012ucf101}, our method has achieved significantly improved accuracy over previous works \cite{vondrick2016generating,saito2016temporal}.

\vspace{-3mm}
\section{Related Work}
\vspace{-1mm}
Image generation by using Generative Adversarial Networks (GANs) \cite{goodfellow2014generative,radford2015unsupervised} has become increasingly popular. For example, in exploiting GAN, Pix2Pix \cite{isola2016image} has succeeded in converting images with the same edges between input and target by using a U-net \cite{ronneberger2015u}. For indoor-scene image generation, Style and Structure GAN (${\rm S^{2}}$GAN) \cite{wang2016generative} focuses on the basic principles by which indoor scene images have a 3D structure in the original world and have texture/style on their surfaces. ${\rm S^{2}}$GAN first generates a 3D surface normal map and then a 2D RGB image, given the generated surface as the condition. In ${\rm S^{2}}$GAN, it is important to consider fundamental information of target domains for generation. Our proposed model is based on these generative models \cite{goodfellow2014generative,radford2015unsupervised,wang2016generative,isola2016image}.

%%%(2段落目)　future frame prediction
Future frame prediction \cite{oh2015action,mathieu2015deep,goroshin2015learning,srivastava2015unsupervised,ranzato2014video,finn2016unsupervised,lotter2016deep,villegas2017decomposing,villegas2017learning,xue2016visual} and future optical flow prediction \cite{walker2015dense,walker2016uncertain} have shown impressive results. These studies are also related to our work in terms of video generation. However, although some methods \cite{mathieu2015deep,villegas2017decomposing,villegas2017learning} exploits adversarial learning, our task completely differs from future frame prediction as follows. 
First, while future frame prediction gives the current frame or even the past frames as a condition, our proposed model generates a video only from Gaussian noise. Moreover, while future frame prediction has a definite ground truth as the generating target, our model discriminates a video based on whether it is real. Therefore, our task can be considered as completely different and a more difficult task than future frame prediction.

%%%(３段落目)　動画認識の話
We make use of an action recognition knowledge in video generation. In action recognition, Two-stream \cite{simonyan2014two} has shown the importance of optical flow. Two-stream learns texture and motion information separately with two networks: RGB-stream and optical-flow-stream. Many later works on action recognition \cite{sun2015human,feichtenhofer2016convolutional,wang2016temporal,zolfaghari2017chained,I3D} employ this idea of  splitting videos into orthogonal information. Thus, it turns out that optical flow is important information for action recognition. In this study, we show the importance of optical flow in video generation.
% In action recognition, Two-stream \cite{simonyan2014two} has shown the importance of optical flow. Many later works on action recognition \cite{sun2015human,diba2016efficient,wang2016actions,feichtenhofer2016convolutional,wang2016temporal,zolfaghari2017chained,I3D} also improves recognition performance by using both RGB images and optical flow as inputs. 

%%%(4段落目)　動画生成GAN
Video GAN (VGAN) \cite{vondrick2016generating} has succeeded to generate scene-consistent videos by generating the foreground and background separately, which brings the generated videos scene consistency. The networks of VGAN consist of 3D convolutions aiming at learning motion information and appearance information simultaneously. However, since this method aims to capture motion and appearance only with single-stream 3D convolutional networks, the generated videos have problems with either visual appearance or motion. Some generated videos have plausible frames but no movement in the generated video. The other generated videos have movements, but their movements are implausible. Optical flow can solve these problems and provides richer motion information not only in action recognition but also in video generation. In action recognition, 3D convolutional networks only with RGB input show inferior performance to networks with RGB and optical flow inputs \cite{I3D}.

Temporal GAN (TGAN) \cite{saito2016temporal} aims to simplify 3D convolutions in order to split appearance/motion information and train network parameters more efficiently. TGAN first applies 2D convolutions to RGB images several times, and then applies 1D temporal convolutions to activations resulting from the 2D convolutions. However, if consecutive frames are made to have a small resolution after applying several downsamplings, almost no spatial change could be seen in the time direction. Actually, in action recognition, TCL path of ${\rm F_{ST}CN}$ \cite{sun2015human}, which has an architecture similar to that of TGAN, does not show improved performance over the RGB stream of Two-stream, which is trained on RGB images frame by frame, on the action recognition dataset UCF101 \cite{soomro2012ucf101}. Therefore, when TGAN is trained on datasets that contain a variety of scenes, although each frame becomes a realistic still image, different appearance and scenes appear in the video, and the movement and consistency can not be assured.

To generate appearance-and-motion-realistic videos, we generate motion information and appearance information in two separate stages. We first generate optical flow to maintain the consistency of motion, and then provide texture information to the generated optical flow maintaining the consistency of appearance.

\begin{figure*}[t]
\begin{center}
  	\includegraphics[width=0.9\hsize]{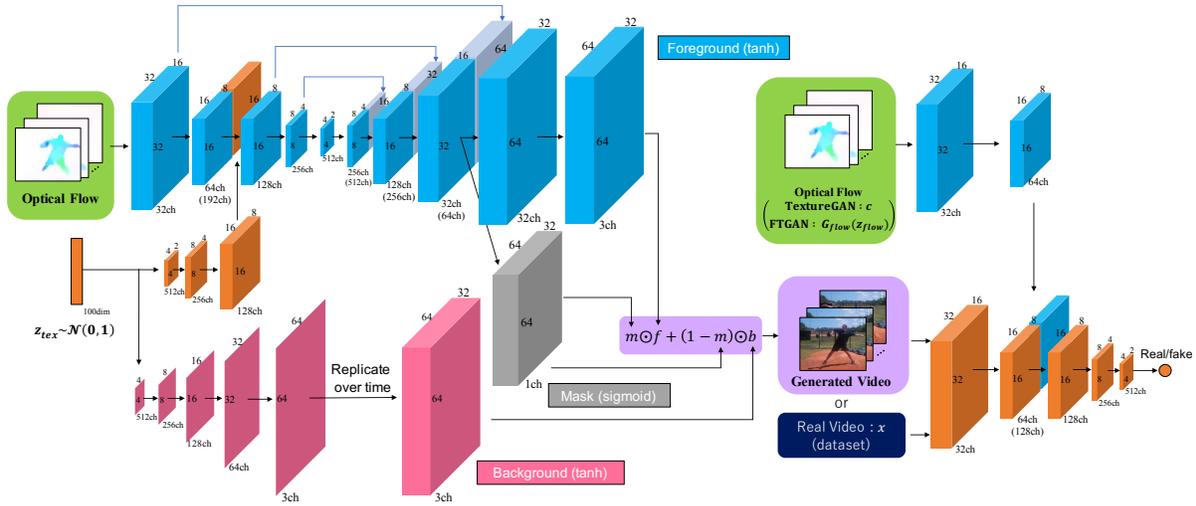}
\end{center}
\vspace{-5mm}
   \caption{{\bf Architecture of Texture GAN:} Given the optical flow video from datasets and $z_{tex}$ as input, Texture Generator learns to generate RGB videos. For foreground generation, we apply 3D convolution to all layers and use skip connection in several layers as a U-net \cite{ronneberger2015u}. For background generation, we apply 2D convolution to all layers. The size of input and output video is 64x64 resolution and 32 frames, which is a little over a second.}
      \label{fig:generator}
     \vspace{-5mm}
\end{figure*}

\vspace{-2mm}
\section{Preliminaries}
\vspace{-1mm}
\subsection{Generative Adversarial Networks}
%\vspace{-2mm}
Generative Adversarial Networks (GANs) \cite{goodfellow2014generative} consist of two networks : Generator (\vector{G}) and Discriminator (\vector{D}). \vector{G} attempts to generate data appearing similar to the given dataset. The input for \vector{G} is a latent variable \vector{z}, which is randomly sampled from distribution $p_{\vector{z}}$ (e.g., a Gaussian distribution). Furthermore, \vector{D} attempts to distinguish between real data and fake data generated from \vector{G}.  A GAN simultaneously updates these two networks, and the objective function is given as follows :  

\begin{equation}
\begin{split}
	\min_{G} \max_{D} V(G,D) &= \mathbb{E}_{\vector{x} \sim p_{data}} [\log(D(\vector{x}))] \\
	&\quad+ \mathbb{E}_{\vector{x} \sim p_{\vector{z}}} [\log (1-D(G(\vector{z})))] 
\end{split}
\end{equation}
\vspace{-5mm}

\subsection{Generative Adversarial Network for Video}
%\vspace{-2mm}
%%%VGANの説明
Generative Adversarial Network for Video (VGAN) \cite{vondrick2016generating} is a video generative network based on the concept of GANs. VGAN also consists of Generator and Discriminator. The generator of VGAN has a mask architecture to separately generate static background and moving foreground: 

%VideoGANのgeneratorの式
\vspace{-3mm}
\begin{equation}
	G(\vector{z}) = m(\vector{z}) \odot f(\vector{z}) + (1 - m(\vector{z})) \odot b(\vector{z})
\end{equation}
\vspace{-3mm}

where $\odot$ represents the element-wise multiplication, $m(\vector{z})$ is a spatiotemporal matrix with each pixel value ranging from 0 to 1; it selects either the foreground $f(\vector{z}$) or the background $b(\vector{z})$ for each pixel $(x,y,t)$. To generate a consistent background, $b(\vector{z})$ produces a spatial static image  replicated over time. During learning, to encourage background image, L1 regularization $\lambda \|m(\vector{z})\|_{1}$ for $\lambda = 0.1$ is added on the mask to the GAN's original objective function.

\section{Method}
Video generation is a difficult task in terms of not only {\it generating natural frame images} but also {\it generating a consistent moving video}. Therefore, to obtain a video comprising plausible motion and realistic frames, we split the generative models into two networks: FlowGAN and TextureGAN. After training each network separately, we conduct joint learning. Figure \ref{fig:system_overview} shows the overview of our FTGAN.

%%%FlowGAN
\begin{figure*}[t]
\begin{center}
  	\includegraphics[width=0.9\hsize]{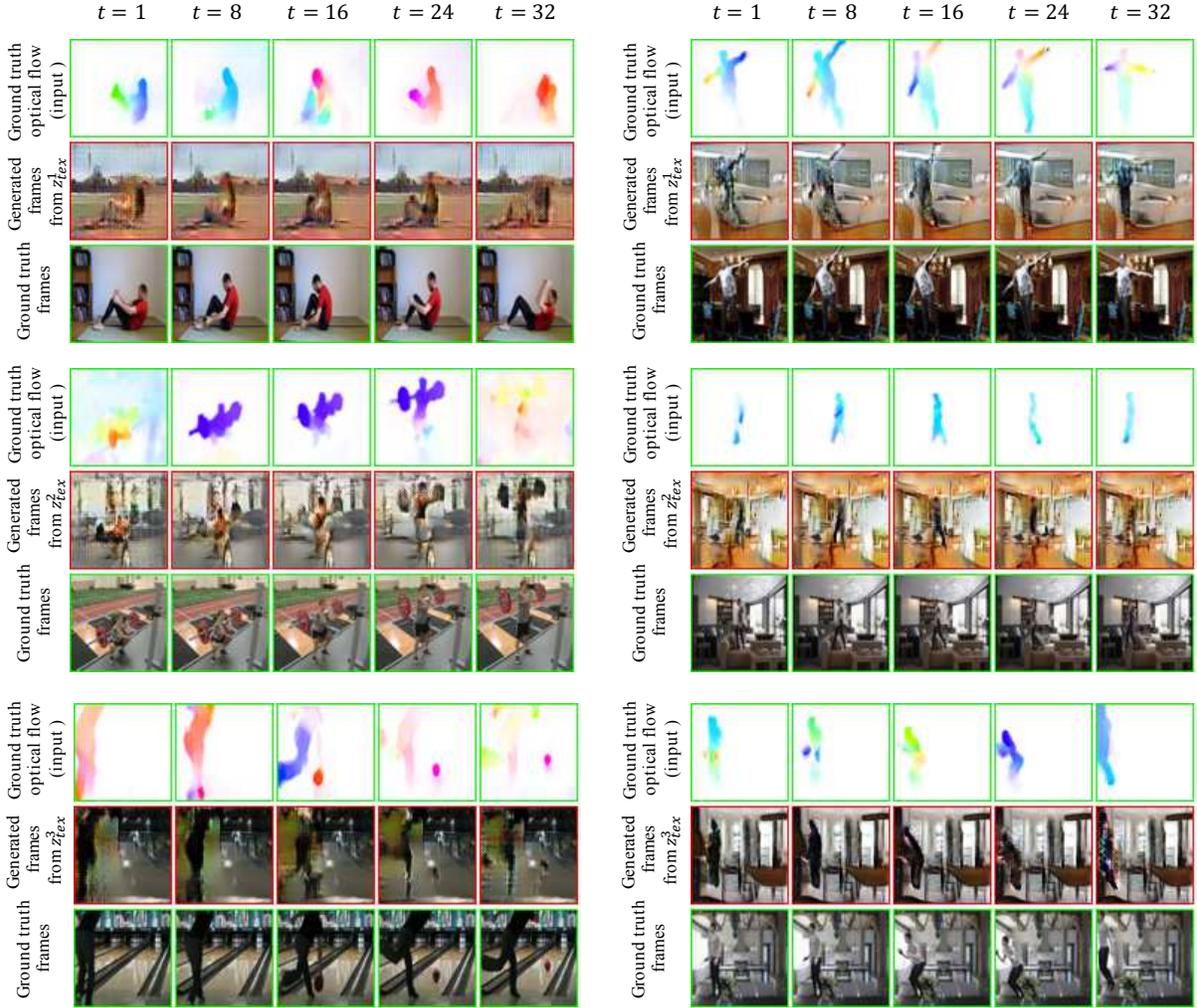}
\end{center}
\vspace{-6mm}
   \caption{Results of TextureGAN conditioned on ground truth optical flow that the network does not observe in the training on two datasets. (Left: Penn Action, Right: SURREAL). The first line in each result shows the input optical flow and the middle line shows the frames generated from the first line's optical flow. The last line shows the ground truth frames corresponding to input optical flow. In the third row of Penn Action, we can see that our model generates not only a person but also the bowling ball. Animated gifs of these results can be seen in the supplemental material. Note that in this figure, $t$ represents the frame number in the videos.}
    \label{fig:texgan}
    \vspace{-2mm}
\end{figure*}

\subsection{FlowGAN: Optical Flow Generation Model}
%\vspace{-2mm}
The architecture of FlowGAN is based on VGAN \cite{vondrick2016generating}. However, considering that the background optical flow should be zero if the camera is fixed, our model does not comprise a background stream in the generator. Instead of learning the background generator, we give the zero matrix as $b$, which is equal to using only the foreground stream of VGAN. 
\begin{equation}
	G_{flow}\left(\vector{z}_{flow} \right) = m\left(\vector{z}_{flow} \right) \odot f\left( \vector{z}_{flow} \right)
\end{equation}

%%%TextureGAN
\vspace{-2mm}
\subsection{TextureGAN: Optical Flow Conditional Video Generation Model}
%\vspace{-2mm}
As one of the simplest ways to generate video using optical flow, we can warp the first image simply along the optical flow. However, the videos generated by such a method tend to be collapsed \cite{villegas2017learning}  especially on relatively long videos. It is more appropriate to fuse image and optical flow after feature-encoding them. Thus we take the following approach. 
As shown in Figure \ref{fig:generator}, our TextureGAN model considers the optical flow and $z_{tex}$ as inputs and outputs a video. The architecture of our generator is based on VGAN and Pix2Pix \cite{isola2016image}. We employ the idea of foreground/background separation from VGAN. For foreground generation, we leverage optical flow which already comproses rough edges of the target video. Thus, we utilize the U-net architecture \cite{ronneberger2015u} as Pix2Pix.

Our Texture Generator $G_{tex}$ is as follows :
\begin{equation}
\begin{split}
	G_{tex}\left(\vector{z}_{tex},\vector{c}\right) &= m\left(\vector{z}_{tex},\vector{c}\right) \odot f\left( \vector{z}_{tex},\vector{c} \right) \\
	&\quad+ \left(1- m\left(\vector{z}_{tex},\vector{c}\right) \right) \odot b\left(\vector{z}_{tex}\right)
\end{split}
\end{equation}
where, $\vector{c}$ is the ground truth optical flow. Note that we provide a {\it ground truth} optical flow to the discriminator as the condition for TextureGAN training.

Then, the loss functions for discriminator and the generator are as follows :
\begin{eqnarray}
\begin{split}
	L_{D_{tex}}\left(\vector{x}, \vector{z}_{tex},\vector{c}\right) &= \log\left(1-D_{tex} \left( G_{tex}\left(\vector{z}_{tex},\vector{c}\right),\vector{c} \right)\right)  \\
	&\quad+  \log\left(D_{tex}\left(\vector{x},\vector{c} \right) \right)   \\
	L_{G_{tex}}\left(\vector{z}_{tex},\vector{c}\right) &= \log\left(D_{tex}\left(G_{tex}\left(\vector{z}_{tex},\vector{c}\right)\right)\right) 
\end{split}
\end{eqnarray}
where, $\vector{x}$ is the ground truth video.

%%%FTGAN
\vspace{-2mm}
\subsection{FTGAN: Both Optical Flow and Video Generation with Joint-Learning}
%\vspace{-1mm}
Figure \ref{fig:system_overview} shows system overview of FTGAN. 
We first train FlowGAN and TextureGAN independently, and then merge these networks through joint learning, during which the generator of the FlowGAN is updated depending on the loss propagated from not only its own discriminator but also the discriminator of TextureGAN. However, the loss from TextureGAN is a supplemental loss for FlowGAN; thus, we set the weighting parameter $\lambda = 0.1$ as ${\rm S^2}$GAN \cite{wang2016generative}. Through this joint learning, we consider that FlowGAN will become able to generate a complementary optical flow suitable for video generation. Note that we use the {\it generated} optical flow as a condition to the discriminator.
\begin{equation}
\begin{split}
	L^{joint}_{G_{flow}}(\vector{z}_{flow}, \vector{z}_{tex}) &= L_{G_{flow}}(\vector{z}_{flow}) \\
	&+ \lambda \cdot L_{G_{tex}} (G_{flow}(\vector{z}_{flow}), \vector{z}_{tex})
\end{split}
\end{equation}
where $\vector{z}_{tex}$ and $\vector{z}_{flow}$ are independent variables.

\begin{figure*}[t]
\begin{center}
  	\includegraphics[width=0.9\hsize]{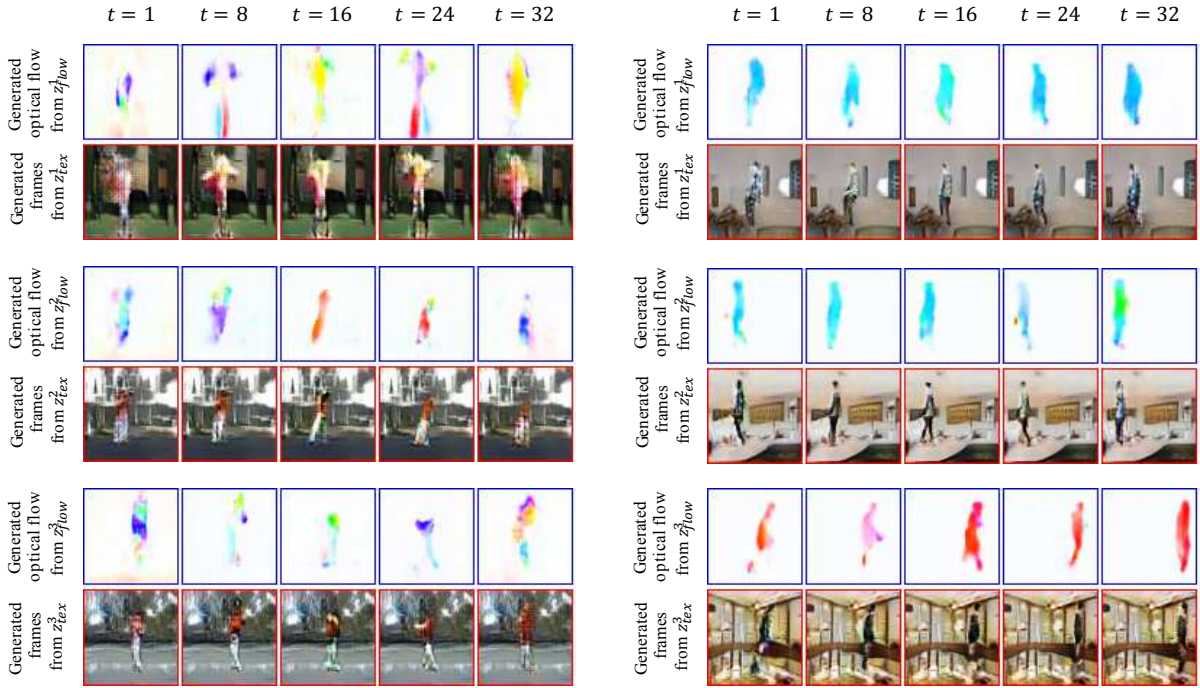}
\end{center}
   \caption{Results of FTGAN generated from $\vector{z}_{flow}$ and $\vector{z}_{tex}$. (Left: Penn Action, Right: SURREAL). Unlike TextureGAN, our networks also generate optical flow images. We can observe that the generated optical flow contains plausible motion. For example, the first row of Penn Action looks like jumping jacks. Although the generated images are clearer when we use the ground truth optical flow than when we also generate optical flow, FTGAN generates the videos in which we can understand what action is being done. For example, in the second row of Penn Action, the person is pitching a ball. In the third row of Penn Action, the person is swinging a baseball bat. In the all results of SURREAL, we can confirm that the optical flow images have reasonable human shape and the generated frames also look realistic frames. Animated gifs of these results can be seen in the supplemental material.}
   \label{fig:ftgan}
   \vspace{-1mm}
\end{figure*}

    \vspace{-2mm}
\subsection{Network Configuration}
%\vspace{-2mm}
First, we train the FlowGAN and TextureGAN independently by using the Adam \cite{kingma2014adam} optimizer with an initial learning rate $\alpha = 0.0002$ and momentum parameter $\beta_{1} = 0.5$. The learning rate is decayed to $1/2$ from its previous value six times during the training. The latent variables $\vector{z}_{tex}$ and $\vector{z}_{flow}$ are Gaussian distributions with 100 dimensions. We set a batch size of $32$. 
Batch normalization \cite{ioffe2015batch} and Rectified Liner Unit (ReLU) activation are applied after every up-sampling convolution, except for the last layer. For down-sampling convolutions, we also apply batch normalization and LeakyReLU \cite{xu2015empirical} to all layers but only apply batch normalization to the first layer. After training them independently, we join both networks and train our FTGAN full-network. Following ${\rm S^2}$GAN, we set a small learning rate $\alpha = 1e-7$ for the FlowGAN and set $\alpha = 1e-6$ for the TextureGAN during the joint learning.

\vspace{-2mm}
%\section{Experiments I: Video Generation}
\section{Experiments}
Evaluation of generative model is difficult due to its lack of appropriate metrics \cite{theis2015note}. Thus, we evaluate our method on generation and recognition tasks following previous works \cite{vondrick2016generating,saito2016temporal}. We first present our experiment of video generation. Next, we present our experiment on classifying actions to explore our model's capability to learn video representations in an unsupervised manner.

%%%Dataset
\vspace{-2mm}
\subsection{Dataset and Settings}
%\vspace{-2mm}
For evaluating video generation, we conduct experiments on two video datasets of human actions. For the real world human video dataset, we use Penn Action \cite{zhang2013actemes}, which has 2326 videos of 15 different classes and 163841 frames. We employ the original train/test split. 
For the Computer Graphics (CG) human video dataset, we use SURREAL \cite{varol17a}, which is made by synthesizing CG humans and LSUN \cite{yu2015lsun} images, and consists of 67582 videos. In the original train/test splits, even test videos have 12538 videos, which contain 1194662 frames. Thus, we use original test videos for training and 1659 subset videos from the original train for testing. 
The TexGAN and FlowGAN are trained in 60000 iterations respectively on each dataset. We conducted joint learning with 10000 iterations on each dataset. For optical flow computation, we use Epic flow \cite{revaud:hal-01142656}. We resize all frames and optical flow images to $76\times76$ resolution, and augment them by cropping them into $64\times64$ resolution images and randomly applying horizontal flips during training. Note that we remove a few videos in each dataset because they have less than 32 frames.%, which corresponds roughly from 1 second to 2 seconds. % More results can be seen in supplemental material.

\begin{table*}[t]
	\small
                \caption{We compare our methods and VGAN with A/B testing. The table shows the percentage of that workers who prefer videos generated from each of our model instead of VGAN.}
  \begin{center}
  \begin{tabular}{c|cc}
    \toprule	
	"Which human video looks more realistic ?"	& SURREAL			&Penn Action \\ \midrule
	Prefer TextureGAN (ours) over VGAN		&	44\%				& 72\%			\\
	Prefer FTGAN (ours) over VGAN			&	54\%				& 58\%			\\ \bottomrule
  \end{tabular}
   \end{center}
   \label{tab:abtest}
   \vspace{-4mm}
\end{table*}

\begin{figure*}[t]
\begin{center}
  	\includegraphics[width=0.9\hsize]{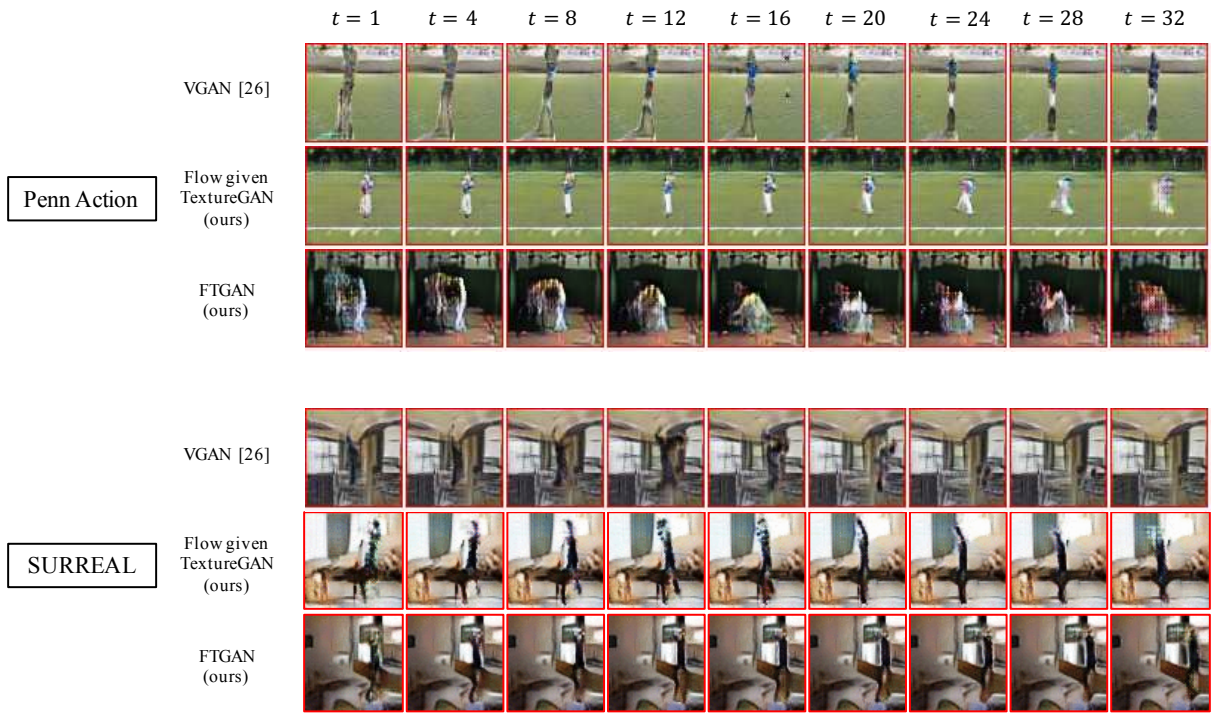}
\end{center}
   \vspace{-4mm}
   \caption{Qualitative comparison of our methods with VGAN. We also show several randomly sampled results of our methods and VGAN in the supplemental material.}
   \label{fig:system}
   \vspace{-5mm}
\end{figure*}

%\newpage
\vspace{-2mm}
%%%TexGAN
\subsection{TextureGAN: Video Generation Results from Ground Truth Optical Flow and $\vector{z}_{tex}$}

Figure \ref{fig:texgan} shows the results of our TextureGAN given the ground truth optical flow on Penn Action and SURREAL. TextureGAN generates videos with plausible motion. Also, on Penn Action, our method generates videos in which not only the motion of the humans appears but also the motion of the objects (e.g. the barbell in the second row , the bowling ball in the last row) appear. This is because the optical flow represents all moving objects as well as people, unlike key-point information, (e.g., human joint positions), which only represent human motion \cite{villegas2017learning}. Moreover, the first row of Penn Action shows that our model generates the video in which a person is doing sit-up in a baseball ground; this combination of appearance and motion information does not exist in this dataset.

\begin{table*}[t]
\small
        \begin{center}
        \caption{Accuracy of unsupervised action classification on UCF101. We train our models with randomly initialized weights. Note that while the first group from the top shows the methods trained with randomly initialized weights, the second group from the top shows the methods pre-trained on 5000 hours video datasets. In addition, while Two-stream \cite{simonyan2014two} is trained in a supervised manner, the other methods, including our proposed methods, are trained in an unsupervised manner.}
  \begin{tabular}{l|ccc}
    \toprule	
Method																	&	Accuracy	\\ \midrule
 Chance																	&	0.9\% 	\\
VGAN + Random Init \cite{vondrick2016generating}									&	36.7\%	\\
TGAN: Image-discriminator + Linear SVM \cite{saito2016temporal}						&	38.6\%	\\
TGAN: Temporal-discriminator + Linear SVM \cite{saito2016temporal}					&	23.3\%	\\
FTGAN {\bf (ours)}: Flow-discriminator + Linear SVM 								&	{\bf 49.5\%}	\\
FTGAN {\bf (ours)}: Texture-discriminator + Linear SVM 								&	{\bf 50.5\%}	\\ 
FTGAN {\bf (ours)}: Flow-discriminator \& Video-discriminator (fusion by Linear SVM) 		&	{\bf 60.9\%}	\\ \midrule
VGAN +  Logistic Reg \cite{vondrick2016generating}									&	49.3\%	\\
VGAN +  Fine Tune \cite{vondrick2016generating}									&	52.1\%	\\ \midrule \midrule
RGB-stream {\small in Two-stream} \cite{simonyan2014two} 											&	73.0\% \\
Flow-stream {\small in Two-stream} \cite{simonyan2014two} 											& 	83.7\% \\
Two-stream (fusion by Linear SVM) \cite{simonyan2014two}  											&	88.0\% \\
    \bottomrule
  \end{tabular}
  \label{tab:ucf101}
\end{center}
\vspace{-7mm}
\end{table*}

%\newpage
%%%FTGAN
%\vspace{-2mm}
\subsection{FTGAN: Video Generation Results from $\vector{z}_{flow}$ and $\vector{z}_{tex}$}
\vspace{-1mm}
In Figure \ref{fig:ftgan}, the results of our FTGAN are presented from only the latent variables $\vector{z}_{flow}$ and $\vector{z}_{tex}$. In this figure, although the results are less clear than the videos generated through the ground truth optical flow, we can observe  plausible moves in the results. For example, in Penn Action, whereas VGAN does not generate videos with plausible motion, our TextureGAN generates a video in which a person is pitching a baseball. For another example, in SURREAL, whereas VGAN generates non-human like frames and disappearing movement, our methods generate human like frames and consistent motion.

Note that looking at the generated videos, it seems like the frames towards the middle of a video are generally better than frames at the beginning and the end. We believe this is caused by the zero-padding in up-sampling convolutions. In each convolutional layer, the beginning and the end frames are partially convolved with zero-padding features. This probably affects the quality of the borders of the generated videos. The spatial borders appear at the edge of each frame, thus they are not usually noticeable. However, the temporal borders appear at the beginning and the end frames (not only edge, but full frames), thus they are noticeable.

%Moreover, we can see that the generated videos from the same latent variable $\vector{z}_{flow}$ have similar moves but are different in the texture, similar to the results of TextrueGAN.

%%%ABtest
\vspace{-2mm}
\subsection{Quantitative Comparison to Previous Study: Human Evaluation}
%\vspace{-2mm}
To evaluate our method quantitatively, it is desirable to allow humans to check the generated results. Thus, we use the Amazon Mechanical Turk (AMT) to compare the results. Table \ref{tab:abtest} shows A/B testing performances of the comparison between our methods and VGAN on AMT. We ask 160 unique workers to indicate which video looks more realistic on 50 videos, and obtained 8000 opinions.

Although TextureGAN shows slightly inferior performance to VGAN on SURREAL, our methods shows improved performance over VGAN. Especially our methods outperform VGAN on Penn Action, which contains more varied motion than SURREAL. We can infer that it is important to divide generation process into motion and appearance as the dataset increases the complexity of motion.

%%%比較
%\vspace{-2mm}
\subsection{Qualitative Comparison to Previous Study: Visualized Results}
We also compare the videos generated from our methods (TextureGAN, FTGAN) with those generated from VGAN, as shown in Figure \ref{fig:system}. In this figure, we introduce examples of videos used for comparison in AMT. On the each dataset, the videos generated by VGAN do not show realistic moves, and the contours of humans are not maintained during the whole video. On the other hand, the videos generated by our methods have plausible motion, and humans are seen in the whole video consistently. We show several randomly sampled gif animations in the supplemental material.

%rebuttal (1-2)
% Looking at the generated videos, it seems like the frames towards the middle of a video are generally better than frames at the beginning and the end. Can authors provide some explanation?
%We believe this is caused by the zero-padding in up-sampling convolutions. In each convolutional layer, the beginning and the end frames are partially convolved with zero-padding features. This probably affects the quality of the borders of the generated videos. The spatial borders appear at the edge of each frame, thus they are not usually noticeable. However, the temporal borders appear at the beginning and the end frames (not only edge, but full frames), thus they are noticeable. Thus, the quality of the beginning and the end frames is worse compared to the middle frames.

%\vspace{-2mm}
%\section{Experiments I\hspace{-.1em}I: Unsupervised Action Classification}
\subsection{Unsupervised Action Classification}

%%%実験概要
As part of the comparison with other GAN methods, we conducted an experiment to investigate the unsupervised feature expression learning capability of FTGAN, as the same way with VGAN and TGAN. Although there are many other unsupervised feature extraction methods, we chose the recent GAN methods to compare the performance of our model since it is also GAN-based.%\textcolor{blue}{Although there are many other unsupervised feature extraction methods, we chose the recent GAN methods to compare the performance of our model since it is also GAN-based.}

%\textcolor{blue}{We also conducted experiments on FTGAN for learning unsupervised representation in videos, similar to VGAN and TGAN.} 
Table \ref{tab:ucf101} shows the action classification performance on UCF101 \cite{soomro2012ucf101}. We train FTGAN on UCF101 with random initialized weights over three splits. As TGAN, we extract the activation of the last layer in the discriminator and learn a Linear SVM. To obtain video features throughout the video, we use a sliding window with overlapping 16 frames to extract features from an entire video as in C3D \cite{tran2014c3d}. For optical flow estimation, we employ the algorithm proposed by Brox et al. \cite{brox2004high} following Two-stream \cite{simonyan2014two}.

%%%結果とその見解
Table \ref{tab:ucf101} shows that both of our discriminators outperform the discriminators of VGAN and TGAN. This improvements suggests that separating information ensures the capture of much richer video characteristics. Moreover, the fusion of both discriminators achieves significantly improved results, in the same way as Two-stream, a supervised method, achieves improved results by fusing the RGB-stream and the Flow-stream. This indicates that each discriminator learns complementary information. In other words, FlowGAN and TextureGAN are considered to learn mainly about motion and appearance information, respectively. The reason why TextureGAN does not comprise the motion information of FlowGAN although TextureGAN uses 3D convolution, could be that the discriminator of TextureGAN receives an optical flow as a condition. Thus, TextureGAN can focus on learning appearance information. 
Note that the results of "VGAN + Logistic Reg" and "VGAN + Fine Tune" have been achieved by 49.3\% and 52.1\% accuracy when pre-trained on 5000 hours video datasets; this is not commensurable with TGAN, "VGAN + Random Init", and our methods. However, the combination of both our discriminators still shows superior performance over them even without pre-training on the 5000 hours video dataset.
%under adverse conditions.

\vspace{-2mm}
\section{Conclusion and Future Work}
In this study, we propose a FTGAN consisting of two GANs; FlowGAN and TextureGAN. Each network deals with complementary information: motion and appearance. Although the generated video has low resolution, is only a few seconds long, our model is able to generate videos successfully with more plausible motion than the other methods (i.e., VGAN). In addition, we have succeeded in capturing significantly improved unsupervised video representations, confirming that each discriminator learns complementary information. Through this paper, we argue that it is important not only for video recognition but also for video generation to focus on a basic underlying principle of video.%; that is, video can be decomposed into motion and appearance.

% future work (rebuttal 1-1, 1-3)
We believe finding other architectures specific for optical flow is a very important future work. It would also be interesting to investigate whether the length of synthesized video plays a factor in the comparison of previous works. The separate treatment of optical flow and texture can potentially reduce the generation space and is expecting better results on longer videos. 

\section{Supplemental}
Our supplemental material is available on 
\href{http://www.mi.t.u-tokyo.ac.jp/assets/publication/hierarchical_video_generation_sup/}{\bf{HERE}}.

\vspace{1mm}
\section{Acknowledgement}
The authors would like to thank Antonio Tejero-de-Pablos, Hiroharu Kato, and Takuhiro Kaneko for providing useful discussions. We also thank Yusuke Mukuta for helping with the optical flow extraction. This work was partially supported by the Ministry of Education, Culture, Sports, Science and Technology (MEXT) as "Seminal Issue on Post-K Computer".
\vspace{-3mm} 
\begin{quote}
\begin{small}
\bibliographystyle{aaai}
\bibliography{paper}
\end{small}
\end{quote}

\end{document}